\documentclass[11pt,a4paper]{article}
\usepackage{amsmath}
\usepackage[hyperref]{emnlp2020}
\usepackage{times}
\usepackage{euscript}
\usepackage{latexsym}
\usepackage{enumitem}

\usepackage{tabularx}
\usepackage{graphicx}       
\usepackage{cleveref}       
\usepackage{booktabs}       
\usepackage{amsfonts}       
\usepackage{nicefrac}       
\usepackage{xcolor}
\usepackage{microtype}

\aclfinalcopy 

\newcolumntype{P}[1]{>{\centering\arraybackslash}p{#1}}

\title{Generating Label Cohesive and Well-Formed Adversarial Claims}
\author{Pepa Atanasova\thanks{~~denotes equal contribution}\hspace{0.1cm} \and Dustin Wright\footnotemark[1]\hspace{0.1cm} \and Isabelle Augenstein \\
  Dept. of Computer Science \\
  University of Copenhagen \\
  Denmark \\
  \texttt{\{pepa|dw|augenstein\}@di.ku.dk}}

\date{}

\begin{document}
\maketitle
\begin{abstract}
Adversarial attacks reveal important vulnerabilities and flaws of trained models. One potent type of attack are \textit{universal adversarial triggers}, which are individual n-grams that, when appended to instances of a class under attack, can trick a model into predicting a target class. However, for inference tasks such as fact checking, these triggers often inadvertently invert the meaning of instances they are inserted in. In addition, such attacks produce semantically nonsensical inputs, as they simply concatenate triggers to existing samples. Here, we investigate how to generate adversarial attacks against fact checking systems that preserve the ground truth meaning and are semantically valid. We extend the HotFlip attack algorithm used for universal trigger generation by jointly minimizing the target class loss of a fact checking model and the entailment class loss of an auxiliary natural language inference model. We then train a conditional language model to generate semantically valid statements, which include the found universal triggers. We find that the generated attacks maintain the directionality and semantic validity of the claim better than previous work.
\end{abstract}

\section{Introduction}
Adversarial examples~\cite{goodfellow2015explaining, szegedy2013intriguing} are deceptive model inputs designed to mislead an ML system into making the wrong prediction. They expose regions of the input space that are outside the training data distribution where the model is unstable. 
It is important to reveal such vulnerabilities and correct for them, especially for tasks such as fact checking (FC). 

\begin{figure}[t]
\centering
\includegraphics[width=0.95 \columnwidth]{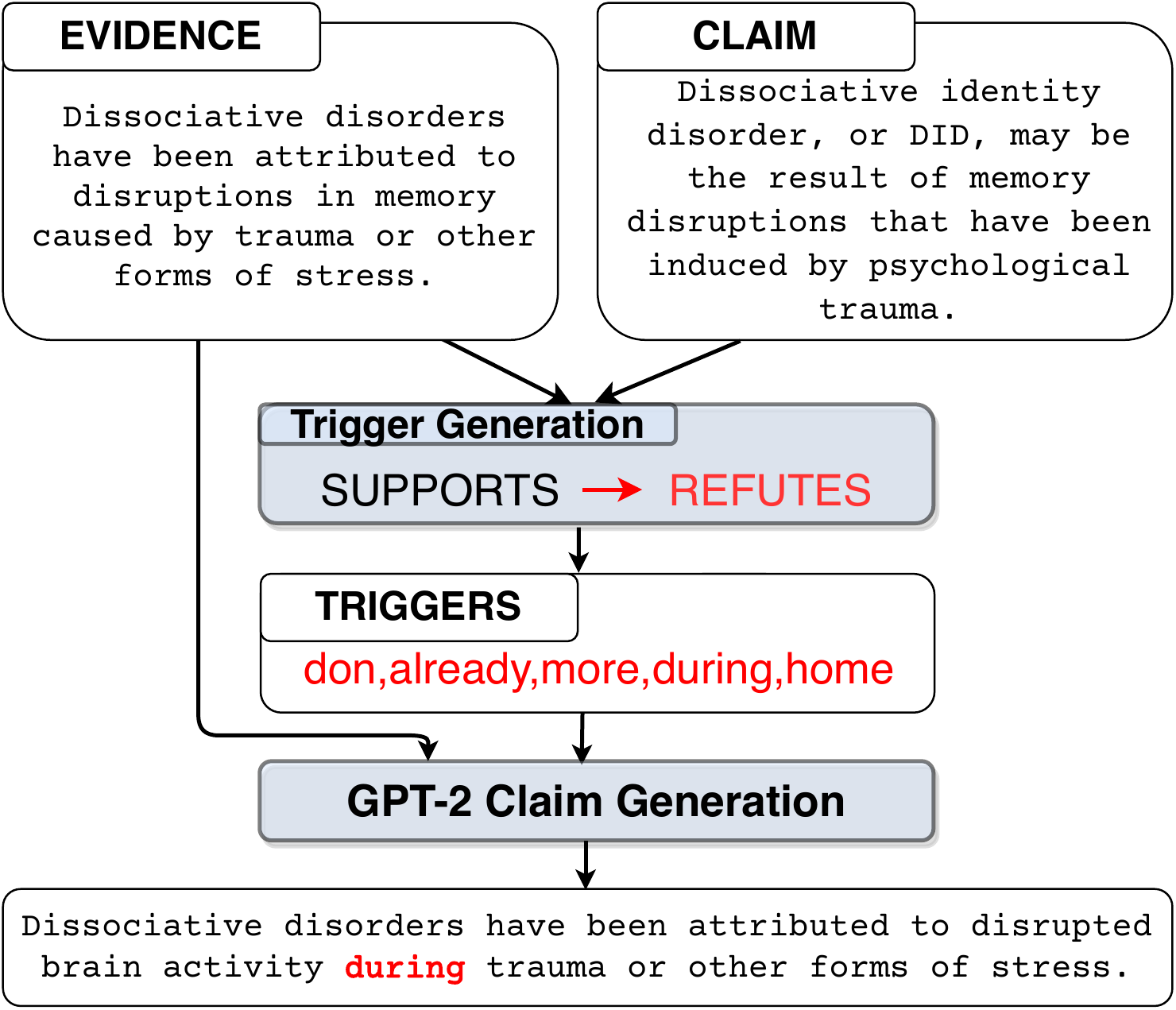}
\caption{High level overview of our method. First, universal triggers are discovered for flipping a source to a target label (e.g. SUPPORTS $\rightarrow$ REFUTES). These triggers are then used to condition the GPT-2 language model to generate novel claims with the original label, including at least one of the found triggers.}
\label{fig:puc}
\end{figure}

In this paper, we explore the vulnerabilities of FC models trained on the FEVER dataset~\cite{thorne2018fever}, where the inference between a claim and evidence text is predicted. We particularly construct \textit{universal adversarial triggers}~\cite{wallace2019universal} -- single n-grams appended to the input text that can shift the prediction of a model from a source class to a target one. Such adversarial examples are of particular concern, as they can apply to a large number of input instances. 

However, we find that the triggers also change the meaning of the claim such that the true label is in fact the target class. For example, when attacking a claim-evidence pair with a `SUPPORTS' label, a common unigram found to be a universal trigger when switching the label to `REFUTES' is `none'. Prepending this token to the claim drastically changes the meaning of the claim such that the new claim is in fact a valid 
`REFUTES' claim as opposed to an adversarial `SUPPORTS' claim. 
Furthermore, we find adversarial examples constructed in this way to be nonsensical, as a new token is simply being attached to an existing claim. 
%

Our \textbf{contributions} are as follows. We \textit{preserve the meaning} of the source text and \textit{improve the semantic validity} of universal adversarial triggers to automatically construct more potent adversarial examples. This is accomplished via: 1) a \textit{novel extension to the HotFlip attack}~\cite{ebrahimi2018hotflip}, where we jointly minimize the target class loss of a FC model and the attacked class loss of a natural language inference model; 2) a \textit{conditional language model} trained using GPT-2~\cite{radford2019language}, which takes 
trigger tokens and a piece of evidence, and generates a semantically coherent new claim containing at least one trigger. 
The resulting triggers maintain potency against a FC model while preserving the original claim label. Moreover, the conditional language model produces semantically coherent adversarial examples containing triggers, on which a FC model performs 23.8\% worse than with the original FEVER claims. The code for the paper is publicly available.\footnote{https://github.com/copenlu/fever-adversarial-attacks}


\section{Related Work}
\subsection{Adversarial Examples}
Adversarial examples for 
NLP systems can be constructed as automatically generated text~\cite{ren2019generating} or perturbations of existing input instances~\cite{jintextfool,ebrahimi2018hotflip}. For a 
detailed literature overview, see~\citet{zhang2019adversarial}.

One potent type of adversarial techniques are universal adversarial attacks~\cite{gao2019universal, wallace2019universal} -- single perturbation changes that can be applied to a large number of input instances and that cause significant performance decreases of the model under attack. 
~\citet{wallace2019universal} find universal adversarial triggers that can change the prediction of the model using the HotFlip algorithm~\cite{ebrahimi2018hotflip}. 

However, for NLI tasks, they also change the meaning of the instance they are appended to, and the prediction of the model remains correct. ~\citet{michel2019evaluation} 
address this by exploring only perturbed instances in the neighborhood of the original one.
Their approach is for instance-dependent attacks, whereas we suggest finding \textit{universal} adversarial triggers that also preserve the original meaning of input instances. 
Another approach to this 
are rule-based perturbations of the input~\cite{ribeiro2018semantically} or imposing adversarial constraints on the produced perturbations~\cite{dia2019semantics}. 
By contrast, we extend the HotFlip method by including an auxiliary Semantic Textual Similarity (STS) objective. We additionally use the extracted universal adversarial triggers to generate adversarial examples with low perplexity.

\subsection{Fact Checking}

Fact checking systems consist of components to identify check-worthy claims \cite{atanasova2018overview,hansen2019neural,wright2020fact}, retrieve and rank evidence documents \cite{conf/emnlp/0001R18,allein2020timeaware}, determine the relationship between claims and evidence documents \cite{bowman2015large,conf/emnlp/AugensteinRVB16,conf/naacl/BalyMGMMN18}, and finally predict the claims' veracity \cite{thorne2018fever,conf/emnlp2019/Augenstein}.
As this is a relatively involved task, models easily overfit to shallow textual patterns, necessitating the need for adversarial examples to evaluate the limits of their performance.

\citet{thorne2019evaluating} are the first to propose hand-crafted adversarial attacks. 
They follow up on this with the FEVER 2.0 
task~\cite{thorne-etal-2019-fever2}, where participants design adversarial attacks for existing FC systems. The first two winning systems~\cite{niewinski-etal-2019-gem} produce claims requiring multi-hop reasoning, which has been shown to be challenging for fact checking models \cite{ostrowski2020multihop}. The other remaining system~\cite{kim-allan-2019-fever} generates adversarial attacks manually. We instead find universal adversarial attacks that can be applied to most existing inputs while markedly decreasing fact checking performance.
\citet{niewinski-etal-2019-gem} additionally feed a pre-trained GPT-2 model with the target label of the instance along with the text for conditional adversarial claim generation. Conditional language generation has also been employed by \citet{keskar2019ctrl} to control the style, content, and the task-specific behavior of a Transformer.

\section{Methods}

\subsection{Models}
We take a RoBERTa~\cite{liu2019roberta} model pretrained with a LM objective and fine-tune it to classify claim-evidence pairs from the FEVER dataset as SUPPORTS, REFUTES, and NOT ENOUGH INFO (NEI). The evidence used is the gold evidence, available for the SUPPORTS and REFUTES classes. For NEI claims, we use the system of \citet{malon2018team} to retrieve evidence sentences. 
To measure the semantic similarity between the claim before and after prepending a trigger, we use a large RoBERTa model fine-tuned on the Semantic Textual Similarity Task.\footnote{https://huggingface.co/SparkBeyond/roberta-large-sts-b} For further details, we refer the reader to \S\ref{sec:appendixA}.

\subsection{Universal Adversarial Triggers Method}
The Universal Adversarial Triggers method is developed to find n-gram trigger tokens $\mathbf{t_{\alpha}}$, which, appended to the original input $x$,  $f(x) = y$, cause the model to predict a target class $\widetilde{y}$ : $f(t_{\alpha}, x) = \widetilde{y}$. In our work, we generate unigram triggers, as generating longer triggers would require additional objectives to later produce well-formed adversarial claims. We start by initializing the triggers with the token `a'. Then, we update the embeddings of the initial trigger tokens $\mathbf{e}_{\alpha}$ with embeddings $\mathbf{e}_{w_i}$ of candidate adversarial trigger tokens $w_i$ that minimize the loss $\mathcal{L}$ for the target class $\widetilde{y}$. Following the HotFlip algorithm, we reduce the brute-force optimization problem using a first-order Taylor approximation around the initial trigger embeddings:
\begin{equation}
\underset{\mathbf{w}_{i} \in \mathcal{V}}{\arg \min }\left[\mathbf{e}_{w_i}-\mathbf{e}_{\alpha}\right]^{\top} \nabla_{\mathbf{e}_{\alpha}} \mathcal{L}
\end{equation}
where $\mathcal{V}$ is the vocabulary of the RoBERTa model and $\nabla_{\mathbf{e}_{\alpha}} \mathcal{L}$ is the average gradient of the task loss accumulated for all batches. This approximation allows for a $\mathcal{O}(|\mathcal{V}|)$ space complexity of the brute-force candidate trigger search.

While HotFlip 
finds universal adversarial triggers that successfully fool the model for many instances, we find that the most potent triggers are often negation words, e.g., `not', `neither', `nowhere'. Such triggers change the meaning of the text, making the prediction of the target class correct. Ideally, adversarial triggers would preserve the original label of the claim. To this end, we propose to include an auxiliary STS model objective when searching for candidate triggers. The additional objective is used to minimize the loss $\mathcal{L'}$ for the maximum similarity score (5 out of 0) between the original claim and the claim with the prepended trigger. Thus, we arrive at the combined optimization problem:
\begin{equation}
\small
\underset{\mathbf{w}_{i} \in \mathcal{V}}{\arg \min }([\mathbf{e}_{w_i}-\mathbf{e}_{\alpha}]^{\top} \nabla_{\mathbf{e}_{\alpha}} \mathcal{L} + [\mathbf{o}_{w_i}-\mathbf{o}_{\alpha}]^{\top} \nabla_{\mathbf{o}_{\alpha}} \mathcal{L'})
\end{equation}
where $\mathbf{o}_w$ is the STS model embedding of word $w$. For the initial trigger token, we use ``[MASK]'' as STS selects candidates from the neighborhood of the initial token.

\subsection{Claim Generation}
\label{sec:claim_generation}
In addition to finding highly potent adversarial triggers, it is also of interest to generate coherent statements containing the triggers. To accomplish this, we use the HuggingFace implementation of the GPT-2 language model~\cite{radford2019language,Wolf2019HuggingFacesTS}, a large transformer-based language model trained on 40GB of text. 
The objective is to generate a coherent claim, which either entails, refutes, or is unrelated a given piece of evidence, while also including trigger words.

The language model is first fine tuned on the FEVER FC corpus with a specific input format. FEVER consists of claims and evidence with the labels \texttt{SUPPORTS}, \texttt{REFUTES}, or \texttt{NOT ENOUGH INFO} (NEI). We first concatenate evidence and claims with a special token. 
Next, to encourage generation of claims with certain tokens, a sequence of tokens separated by commas is prepended to the input. For training, the sequence consists of a single token randomly selected from the original claim, and four random tokens from the vocabulary. 
This encourages the model to only select the one token most likely to form a coherent and correct claim. The final input format is \texttt{[trigger tokens]}\textbar\textbar\texttt{[evidence]}\textbar\textbar\texttt{[claim]}.
Adversarial claims are then generated by providing an initial input of a series of five comma-separated trigger tokens plus evidence, and progressively generating the rest of the sequence. Subsequently, the set of generated claims is pruned to include only those which contain a trigger token, 
and constitute the desired label. The latter is ensured by passing both evidence and claim through an external NLI model trained on SNLI \cite{bowman2015large}. 

\section{Results}
We present results for universal adversarial trigger generation and coherent claim generation. 
Results are measured using the original FC model on claims with added triggers and generated claims (macro F1). We also measure how well the added triggers maintain the claim's original label (semantic similarity score), the perplexity (PPL) of the claims with prepended triggers, and the semantic quality of generated claims (manual annotation). PPL is measured with a pretrained RoBERTa LM.

\subsection{Adversarial Triggers}
Table~\ref{tab:eval} presents the results of applying universal adversarial triggers to claims from the source class.
The top-performing triggers for each direction are found in \S\ref{sec:appendixC}. 
The adversarial method with a single FC objective successfully deteriorates model performance by a margin of 0.264 F1 score overall. The biggest performance decrease is when the adversarial triggers are constructed to flip the predicted class from SUPPORTS to REFUTES. We also find that 8 out of 18 triggers from the top-3 triggers for each direction, are negation words such as  `nothing', `nobody', `neither', `nowhere' (see Table~\ref{tab:evalonetrig} in the appendix). The first of these triggers decreases the performance of the model to 0.014 in F1. While this is a significant performance drop, these triggers also flip the meaning of the text. The latter is again indicated by the decrease of the semantic similarity between the claim before and after prepending a trigger token, which is the largest for the SUPPORTS to REFUTES direction. We hypothesise that the success of the best performing triggers is partly due to the meaning of the text being flipped.

Including the auxiliary STS objective increases the similarity between the claim before and after prepending the trigger for five out of six directions. Moreover, we find that now only one out of the 18 top-3 triggers for each direction are negation words. Intuitively, these adversarial triggers are worse at fooling the FC model as they also have to preserve the label of the original claim. Notably, for the SUPPORTS to REFUTES direction the trigger performance is decreased with a margin of 0.642 compared to the single FC objective.
We conclude that including the STS objective for generating Universal Adversarial triggers helps to preserve semantic similarity with the original claim, but also makes it harder to both find triggers preserving the label of the claim while substantially decreasing the performance of the model.

\begin{table}[t]
\small
\centering
\begin{tabular}{l@{\hspace{1.2\tabcolsep}}l@{\hspace{1.2\tabcolsep}}l@{\hspace{1.2\tabcolsep}}l}
\toprule
\textbf{Class} & \textbf{F1} & \textbf{STS} & \textbf{PPL}\\ \midrule
\multicolumn{4}{c}{\bf No Triggers} \\
All & .866 & 5.139 & 11.92 ($\pm$45.92) \\
S & .938 & 5.130 & 12.22 ($\pm$40.34) \\
R & .846 & 5.139 &  12.14 ($\pm$37.70) \\
NEI & .817 & 5.147 & 14.29 ($\pm$84.45) \\
\midrule
\multicolumn{4}{c}{\bf FC Objective} \\
All & .602 ($\pm$.289) & 4.586 ($\pm$.328) & 12.96 ($\pm$55.37) \\
S$\rightarrow$R & .060 ($\pm$.034) & 4.270 ($\pm$.295) & 12.44 ($\pm$41.74) \\
S$\rightarrow$NEI & .611 ($\pm$.360) & 4.502 ($\pm$.473) & 12.75 ($\pm$40.50) \\
R$\rightarrow$S & .749 ($\pm$.027) & 4.738 ($\pm$.052) & 11.91 ($\pm$36.53) \\
R$\rightarrow$NEI & .715 ($\pm$.026) & 4.795 ($\pm$.094) & 11.77 ($\pm$36.98) \\
NEI$\rightarrow$R & .685 ($\pm$.030) & 4.378 ($\pm$.232) & 14.20 ($\pm$83.32) \\
NEI$\rightarrow$S & .793 ($\pm$.054) & 4.832 ($\pm$.146) & 14.72 ($\pm$93.15) \\
\midrule
\multicolumn{4}{c}{\bf FC+STS Objectives} \\
All & .763 ($\pm$.123) & 4.786 ($\pm$.156) & 12.97 ($\pm$58.30) \\
S$\rightarrow$R & .702 ($\pm$.237) & 4.629 ($\pm$.186) & 12.62 ($\pm$41.91) \\
S$\rightarrow$NEI & .717 ($\pm$.161) & 4.722 ($\pm$.152) & 12.41 ($\pm$39.66) \\
R$\rightarrow$S & .778 ($\pm$.010) & 4.814 ($\pm$.141) & 11.93 ($\pm$37.04) \\
R$\rightarrow$NEI & .779 ($\pm$.009) & 4.855 ($\pm$.098) & 12.20 ($\pm$37.67) \\
NEI$\rightarrow$R & .780 ($\pm$.078) & 4.894 ($\pm$.115) & 15.27 ($\pm$111.2) \\
NEI$\rightarrow$S & .821 ($\pm$.008) & 4.800 ($\pm$.085) & 13.42 ($\pm$82.30) \\
\bottomrule
\end{tabular}
\caption{Universal Adversarial Trigger method performance. Triggers are generated given claims from a source class to fool the classifier to predict a target class (column \textit{Class}, with SUPPORTS (S), REFUTES (R), NEI). 
The results are averaged over the top 10 triggers.}
\label{tab:eval}
\end{table}

\begin{table*}[!ht]
\fontsize{9}{9}\selectfont
\renewcommand{\arraystretch}{1.4}
\centering
\begin{tabular}{p{6cm} P{4.2cm} p{4.6cm}}
\toprule
\textbf{Evidence} & \textbf{Triggers} & \textbf{Generated Claim}\\ 
\midrule

\multicolumn{3}{c}{\bf SUPPORTS Claims} \\
Since the 19th century, some Romani have also migrated to the Americas. & don,already,more,\textcolor{red}{during},home & Romani have moved to the Americas \textcolor{red}{during} the 19th century. \\

Cyprus is a major tourist destination in the Mediterranean. & foreign,biggest,\textcolor{red}{major},every, friends & Cyprus is a \textcolor{red}{major} tourist destination. \\

The first Nobel Prize in Chemistry was awarded in 1901 to Jacobus Henricus van't Hoff, of the Netherlands, ``for his discovery of the laws of chemical dynamics and osmotic pressure in solutions.'' & later,\textcolor{red}{already},quite,altern, whereas & Henricus Van't Hoff was \textcolor{red}{already} awarded the Nobel Prize. \\
\midrule
\multicolumn{3}{c}{\bf REFUTES Claims} \\

California Attorney General Kamala Harris defeated Sanchez , 61.6\% to 38.4\%. & phys,incarn,\textcolor{red}{not},occasionally,
something & Kamala Harris did \textcolor{red}{not} defeat Sanchez, 61.6\% to 38.4\%. \\

Uganda is in the African Great Lakes region. & unless,endorsed,picks,pref, \textcolor{red}{against} & Uganda is \textcolor{red}{against} the African Great Lakes region. \\

Times Higher Education World University Rankings is an annual publication of university rankings by Times Higher Education (THE) magazine. & interested,reward,visit, \textcolor{red}{consumer},conclusion & Times Higher Education World University Rankings is a \textcolor{red}{consumer} magazine. \\

\midrule
\multicolumn{3}{c}{\bf NOT ENOUGH INFO Claims} \\
The KGB was a military service and was governed by army laws and regulations, similar to the Soviet Army or MVD Internal Troops. & nowhere,\textcolor{red}{only},none,no,nothing & The KGB was \textcolor{red}{only} controlled by a military service. \\

The series revolves around Frank Castle, who uses lethal methods to fight crime as the vigilante ``the Punisher'', with Jon Bernthal reprising the role from Daredevil. & says,said,\textcolor{red}{take},say,is & \textcolor{red}{Take} Me High is about Frank Castle's use of lethal techniques to fight crime. \\

The Suite Life of Zack \& Cody is an American sitcom created by Danny Kallis and Jim Geoghan. & whilst,interest,applic,\textcolor{red}{someone}, nevertheless & The Suite Life of Zack \& Cody was created by \textcolor{red}{someone} who never had the chance to work in television. \\
\bottomrule
\end{tabular}
\caption{Examples of generated adversarial claims. These are all claims which the FC model incorrectly classified.}
\label{tab:generation_examples}
\end{table*}

\subsection{Generation}
We use the method described in \S\ref{sec:claim_generation} to generate 156 claims using triggers found with the additional STS objective, and 156 claims without. 52 claims are generated for each class (26 flipping to one class, 26 flipping to the other). A different GPT-2 model is trained to generate claims for each specific class, with triggers specific to attacking that class used as input. The generated claims are annotated manually (see \S\ref{app:B3} for the procedure). The overall average claim quality is 4.48, indicating that most generated statements are highly semantically coherent. The macro F1 of the generative model w.r.t. the intended label is 58.9 overall. For the model without the STS objective, the macro F1 is 56.6, and for the model with the STS objective, it is 60.7, meaning that using triggers found with the STS objective helps the generated claims to retain their intended label.

We measure the performance of the original FC model on generated claims (\autoref{tab:generation_eval}). We compare between using triggers that are generated with the STS objective (Ex2) and without (Ex1). In both cases, the adversarial claims effectively fool the FC model, which performs 38.4\% worse and 23.8\% worse on Ex1 and Ex2, respectively.  Additionally, the overall sentence quality increases when the triggers are found with the STS objective (Ex2). The FC model's performance is higher on claims using triggers generated with the STS objective but still significantly worse than on the original claims. We provide examples of generated claims with their evidence in \autoref{tab:generation_examples}.
\begin{table}
\fontsize{10}{10}\selectfont
\centering
\begin{tabular}{lccc}
\toprule
\textbf{Target} & \textbf{F1} & \textbf{Avg Quality} & \textbf{\# Examples}\\ \midrule
\multicolumn{4}{c}{\bf FC Objective} \\
Overall& 0.534& 4.33&156\\
SUPPORTS& 0.486& 4.79& 39\\
REFUTES& 0.494& 4.70&32\\
NEI& 0.621& 3.98 &85\\
\midrule
\multicolumn{4}{c}{\bf FC+STS Objectives} \\
Overall& 0.635& 4.63&156\\
SUPPORTS& 0.617& 4.77&67\\
REFUTES& 0.642& 4.68&28\\
NEI& 0.647& 4.44&61\\
\bottomrule
\end{tabular}
\caption{FC performance for generated claims.}
\label{tab:generation_eval}
\end{table}

Comparing FC performance with our generated claims vs. those from the development set of adversarial claims from the FEVER shared task 
, we see similar drops in performance (0.600 and 0.644 macro F1, respectively). While the adversarial triggers from FEVER cause a larger performance drop, they were manually selected to meet the label coherence and grammatical correctness requirements. Conversely, we automatically generate claims that meet these requirements.

\section{Conclusion}
We present a method for automatically generating highly potent, well-formed, label cohesive claims for FC. 
We improve upon previous work on universal adversarial triggers by determining how to construct valid claims containing a trigger word. 
Our method is fully automatic, whereas previous work on generating claims for fact checking is generally rule-based or requires manual intervention. As FC is only one test bed for adversarial attacks, it would be interesting to test this method on other NLP tasks requiring semantic understanding such as question answering 
to better understand shortcomings of models. 

\section*{Acknowledgements}
$\begin{array}{l}\includegraphics[width=1cm]{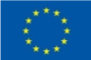} \end{array}$ This project has received funding from the European Union's Horizon 2020 research and innovation programme under the Marie Sk\l{}odowska-Curie grant agreement No 801199.

\clearpage

\bibliography{anthology,acl2020}
\bibliographystyle{acl_natbib}

\appendix
\section{Appendices}
\subsection{Implementation Details}
\label{sec:appendixA}
\textbf{Models}. The RoBERTa FC model (125M parameters) is fine-tuned with a batch size of 8, learning rate of 2e-5 and for a total of 4 epochs, where the epoch with the best performance is saved. We used the implementation provided by HuggingFace library. We performed a grid hyper-parameter search for the learning rate between the values 1e-5, 2e-5, and 3e-5. The average time for training a model with one set of hyperparameters is 155 minutes ($\pm3$). The average accuracy over the different hyperparameter runs is 0.862($\pm$ 0.005) F1 score on the validation set.

For the models that measure the perplexity and the semantical similarity we use the pretrained models provided by HuggingFace-- RoBERTa large model (125M parameters) fine tuned on the STS-b task and RoBERTa base model (355M parameters) pretrained on a LM objective.

We used the HuggingFace implementation of the small GPT-2 model, which consists of 124,439,808 parameters and is fine-tuned with a batch size of 4, learning rate of 3e-5, and for a total of 20 epochs. We perform early stopping on the loss of the model on a set of validation data. The average validation loss is 0.910. The average runtime for training one of the models is 31 hours and 28 minutes.

We note that, the intermediate models used in this work and described in this section, are trained on large relatively general-purpose datasets. While, they can make some mistakes, they work well enough and using them, we don't have to rely on additional human annotations for the intermediate task.

\textbf{Adversarial Triggers.} The adversarial triggers are generated based on instances from the validation set. We run the algorithm for three epochs to allow for the adversarial triggers to converge. At each epoch the initial trigger is updated with the best performing trigger for the epoch (according to the loss of the FC or FC+STS objective). At the last step, we select only the top 10 triggers and remove any that have a negative loss. We choose the top 10 triggers as those are the most potent ones, adding more than top ten of the triggers preserves the same tendencies in the results, but smooths them as further down the list of adversarial attacks, the triggers do not decrease the performance of the model substantially. This is also supported by related literature~\cite{wallace2019universal}, where only the top few triggers are selected.

The adversarial triggers method is run for 28.75 ($\pm$ 1.47) minutes for with the FC objective and 168.6($\pm$ 28.44) minutes for the FC+STS objective. We perform the trigger generation with a batch size of four. We additionally normalize the loss for each objective to be in the range [0,1] and also re-weight the losses with a wieht of 0.6 for the FC loss and a weight of 0.4 for the SST loss as when generated with an equal weight, the SST loss tends to preserve the same initial token in all epochs.

\textbf{Datasets.} 
The datasets used for training the FC model consist of 161,249 SUPPORTS, 60,227 REFUTES, and 69,885 NEI claims for the training split; 6,207 SUPPORTS, 6,235 REFUTES, and 6,554 NEI claims for the dev set; 6,291 SUPPORTS, 5,992 REFUTES, and 6522 NEI claims. The evidence for each claim is the gold evidence provided from the FEVER dataset, which is available for REFUTES and SUPPORTS claims. When there is more than one annotation of different evidence sentences for an instance, we include them as separate instances in the datasets. For NEI claims, we use the system of \citet{malon2018team} to retrieve evidence sentences. 

\subsection{Top Adversarial Triggers}
Table~\ref{tab:evalonetrig} presents the top adversarial triggers for each direction found with the Universal Adversarial Triggers method. It offers an additional way of estimating the effectiveness of the STS objective by comparing the number of negation words generated by the basic model (8) and the STS objective (2) in the top-3 triggers for each direction.
\label{sec:appendixC}
\begin{table*}[t]
\centering
\begin{tabular}{l@{\hspace{1.2\tabcolsep}}l@{\hspace{1.2\tabcolsep}}l@{\hspace{1.2\tabcolsep}}l@{\hspace{1.2\tabcolsep}}l}
\toprule
\textbf{Class} & \textbf{Trigger} & \textbf{F1} & \textbf{STS} & \textbf{PPL}\\ \midrule
\multicolumn{5}{c}{\bf FC Objective} \\
S$\rightarrow$R & only &  0.014 &  4.628 &  11.660 (36.191) \\
S$\rightarrow$R & nothing &  0.017 &  4.286 &  13.109 (56.882) \\
S$\rightarrow$R & nobody &  0.036 &  4.167 &  12.784 (37.390) \\
S$\rightarrow$NEI & neither &   0.047 &  3.901 &  11.509 (31.413) \\
S$\rightarrow$NEI & none &  0.071 &  4.016 &  13.136 (39.894) \\
S$\rightarrow$NEI & Neither &  0.155 &  3.641 &  11.957 (44.274) \\
R$\rightarrow$S & some &  0.687 &  4.694 &  11.902 (33.348) \\
R$\rightarrow$S & Sometimes &  0.724 &  4.785 &  10.813 (32.058) \\
R$\rightarrow$S & Some &  0.743 &  4.713 &  11.477 (37.243) \\
R$\rightarrow$NEI & recommended &  0.658 &  4.944 &  12.658 (36.658) \\
R$\rightarrow$NEI & Recommend &  0.686 &  4.789 &  10.854 (32.432) \\
R$\rightarrow$NEI & Supported &  0.710 &  4.739 &  11.972 (40.267) \\
NEI$\rightarrow$R & Only &  0.624 &   4.668 &  12.939 (57.666) \\
NEI$\rightarrow$R & nothing &  0.638 &  4.476 &   11.481 (48.781) \\
NEI$\rightarrow$R & nobody & 0.678 &  4.361 &  16.345 (111.60) \\
NEI$\rightarrow$S & nothing &  0.638 &  4.476 &  18.070 (181.85) \\
NEI$\rightarrow$S & existed &  0.800 &  4.950  &  15.552 (79.823) \\
NEI$\rightarrow$S & area &  0.808 &  4.834  &  13.857 (93.295) \\

\midrule
\multicolumn{5}{c}{\bf FC+STS Objectives} \\
S$\rightarrow$R & never & 0.048 & 4.267 & 12.745 (50.272) \\
S$\rightarrow$R & every & 0.637 & 4.612 & 13.714 (51.244) \\
S$\rightarrow$R & didn & 0.719 & 4.986 & 12.416 (41.080) \\
S$\rightarrow$NEI & always  & 0.299 &  4.774 &  11.906 (35.686) \\
S$\rightarrow$NEI & every & 0.637 & 4.612 & 12.222 (38.440) \\
S$\rightarrow$NEI & investors & 0.696 & 4.920 & 12.920 (42.567) \\
R$\rightarrow$S & over &  0.761 &  4.741 &  12.139 (33.611) \\
R$\rightarrow$S & about &  0.765 &   4.826 &  12.052 (37.677) \\
R$\rightarrow$S & her &   0.774 &   4.513 &   12.624 (41.350) \\
R$\rightarrow$NEI & top &  0.757 &  4.762 &  12.787 (39.418) \\
R$\rightarrow$NEI & also &   0.770 &   5.034 &   11.751 (35.670) \\
R$\rightarrow$NEI & when &   0.776 &   4.843 &   12.444 (37.658) \\
NEI$\rightarrow$R & only &  0.562 &  4.677 &  14.372 (83.059) \\
NEI$\rightarrow$R & there &   0.764 & 4.846 &    11.574 (42.949) \\
NEI$\rightarrow$R & just &   0.786 & 4.916 &   16.879 (135.73) \\
NEI$\rightarrow$S & of&   0.802 & 4.917 &  11.844 (55.871) \\
NEI$\rightarrow$S & is &   0.815 & 4.931 & 17.507 (178.55) \\
NEI$\rightarrow$S & A &   0.818 & 4.897 & 12.526 (67.880) \\

\bottomrule
\end{tabular}
\caption{Top-3 triggers found with the Universal Adversarial Triggers methods. The triggers are generated given claims from a source class (column \textit{Class}), so that the classifier is fooled to predict a different target class. The classes are SUPPORTS (S), REFUTES (R), NOT ENOUGH INFO (NEI).}
\label{tab:evalonetrig}
\end{table*}

\section{Supplemental Material}
\label{sec:supplemental}

\subsection{Computing Infrastructure}
All experiments were run on a shared cluster. Requested jobs consisted of 16GB of RAM and 4 Intel Xeon Silver 4110 CPUs. We used two NVIDIA Titan RTX GPUs with 12GB of RAM for training GPT-2 and one NVIDIA Titan X GPU with 8GB of RAM for training the FC models and finding the universal adversarial triggers.

\subsection{Evaluation Metrics}
The primary evaluation metric used was macro-F1 score. We used the sklearn implementation of \texttt{precision\_recall\_fscore\_support}, which can be found here: \url{https://scikit-learn.org/stable/modules/generated/sklearn.metrics.precision_recall_fscore_support.html}. Briefly:
\begin{equation*}
   p = \frac{tp}{tp + fp} 
\end{equation*}
\begin{equation*}
   r = \frac{tp}{tp + fn} 
\end{equation*}
\begin{equation*}
   F1 = \frac{2*p*r}{p+r} 
\end{equation*}
where $tp$ are true positives, $fp$ are false positives, and $fn$ are false negatives.

\subsection{Manual Evaluation}
\label{app:B3}
After generating the claims, two independent annotators label the overall claim quality (score of 1-5) and the true label for the claim. The inter-annotator agreement for the quality label using Krippendorff's alpha is 0.54 for the quality score and 0.38 for the claim label. Given this, we take the average of the two annotator's scores for the final quality score and have a third expert annotator examine and select the best label for each contested claim label.

\end{document}